\documentclass[11pt]{article}

\usepackage[letterpaper,margin=1in]{geometry}
\usepackage[T1]{fontenc}
\usepackage[utf8]{inputenc}
\usepackage{mathptmx}          
\usepackage{microtype}

\usepackage{graphicx}
\usepackage{booktabs}
\usepackage{multirow}
\usepackage{amsmath,amssymb}
\usepackage{caption}
\usepackage[hyphens]{url}
\usepackage[round]{natbib}     
\usepackage[colorlinks=true,linkcolor=blue!50!black,citecolor=blue!50!black,urlcolor=blue!50!black]{hyperref}
\usepackage{xcolor}
\let\cite\citep                
\newcommand{\nullc}{\varnothing}

\title{Robust LLM-Generated Text Detection through\\Matched Conditional-Response Probing}
\author{
  Junfeng Hao\textsuperscript{1},
  Wenxian Wang\textsuperscript{2},
  Tao Leng\textsuperscript{4},
  Xingshu Chen\textsuperscript{1,2,3},
  Hao Ren\textsuperscript{1},
  Haizhou Wang\textsuperscript{1}\thanks{Corresponding author.}\\[0.6em]
  \textsuperscript{1}School of Cyber Science and Engineering, Sichuan University\\
  \textsuperscript{2}Cyber Science Research Institute, Sichuan University\\
  \textsuperscript{3}Key Laboratory of Data Protection and Intelligent Management, Sichuan University\\
  \textsuperscript{4}Intelligent Policing Key Laboratory of Sichuan Province, Sichuan Police College
}
\date{}

\begin{document}

\maketitle

\begin{abstract}
Reliable detection of large language model (LLM)-generated text must remain effective under meaning-preserving text revisions, such as rewriting, polishing, and expansion. In practice, LLM outputs are often rewritten, polished, expanded, translated, or lightly edited before publication, weakening the lexical, syntactic, and local discourse cues used by many existing detectors. We introduce \textbf{M}atched \textbf{C}onditional-\textbf{R}esponse \textbf{P}robing (\textbf{MCRP}), a frozen-LLM probing framework that measures how a candidate text responds when a semantically matched condition is introduced, removed, or replaced. For each candidate, the frozen probe LLM first generates a short semantic condition and then compares token-level losses and prompt--target attention measurements under matched, null, and control conditions. It aggregates bidirectional conditioned alignment, counterfactual sensitivity, and internal attention coupling into a probe representation, on which it trains a lightweight supervised random-forest classifier while keeping the probe LLM frozen. Across standard detection, robustness to meaning-preserving text revisions, multilingual, and multidomain evaluations, MCRP achieves consistently strong performance. Under rewriting, polishing, and expansion, it reaches 93.97\% F1 and 96.24\% ROC-AUC, outperforming the strongest evaluated baseline by 19.50 and 25.77 percentage points, respectively. Component ablations show that the response feature groups make complementary contributions to predictive performance. In the condition-control study, matched conditions consistently outperform random, mismatched, and corrupted alternatives, demonstrating a measurable incremental benefit from semantic matching.
\end{abstract}

\section{Introduction}

Large language models are now used for news summaries, product reviews, question answering, academic drafts, social media posts, and writing assistance, making the detection of LLM-generated text important to trustworthy AI and information governance \cite{sadasivan2025,solaiman2019,yang2023survey,wu2024survey}. Real outputs, however, rarely remain in their original form. Before publication or submission, they may be rewritten, polished, expanded, translated, or lightly edited \cite{pu2023zero,koike2024outfox,pedrotti2025stress,dugan2024raid,wang2024m4gt,cornelius2024bust,wang2024semeval,li2024mage,he2024mgtbench,macko2023multitude}. Such operations change word choice, syntax, local discourse organization, and stylistic regularities while often retaining the core topic and information structure. Consequently, post-processing can weaken surface-level evidence without necessarily removing every relation between the text and an LLM-conditioned generation process \cite{krishna2023paraphrasing}.

Most existing detectors assign a score or label to an observed candidate using likelihood, perplexity, token rank, probability curvature, model discrepancy, rewriting behavior, repair effort, stylistic features, or supervised representations \cite{gehrmann2019gltr,su2023detectllm,venkatraman2024gptwho,verma2024ghostbuster,bao2024fast,mitchell2023detectgpt,hans2024binoculars,chen2025imbd,kirchenbauer2023watermark,liu2024adaptive,wang2023seqxgpt,wu2023llmdet,liu2023coco,corizzo2023onegpt,lee2024adversarial,zhou2024cnet,hou2024ksemstamp}. These forms of evidence are effective in many unchanged, in-distribution, or closed-set settings. Yet their behavior can change after paraphrasing, polishing, translation, domain shift, or generator shift \cite{pu2023zero,dugan2024raid,yu2025evobench,fraser2025factors}. This motivates a complementary question: rather than asking only whether a text appears LLM-generated under a static score, can we test whether its response changes systematically when a semantically matched condition is introduced, removed, or replaced?

\begin{figure}[t]
\centering
\includegraphics[width=\textwidth]{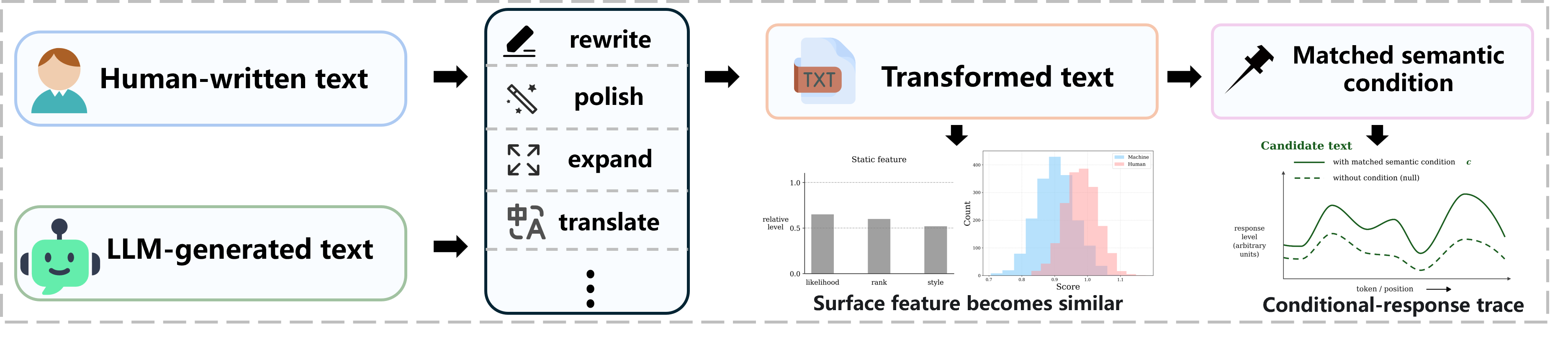}
\caption{Motivation under meaning-preserving text revisions. Human-written and LLM-generated texts may both be paraphrased, polished, expanded, or translated, making static surface evidence less separable. We instead reintroduce a semantic condition matched to the transformed candidate and measure its conditional-response profile.}
\label{fig:motivation}
\end{figure}

Figure~\ref{fig:motivation} illustrates the test performed by MCRP. Given a candidate text ($x$), we first use the frozen probe model ($f_\theta$) to generate a short title ($c$) summarizing its main topic. The same model then processes ($x$) under two contexts: a matched context containing ($c$) and a null context containing only the completion instruction. For each target token, we measure how the matched condition changes its prediction loss and how much attention it receives from the target sequence; we also measure how well the candidate predicts the condition. Our hypothesis is that LLM-generated text is often organized as an expansion of a compact instruction or topic specification. Consequently, even after rewriting or polishing, introducing a candidate-matched condition may produce a systematically different loss-reduction and attention profile for machine- and human-written text. This is a testable difference between the two classes, not an attempt to recover the original prompt or infer the true causal generation process.

Based on this view, we propose Matched Conditional-Response Probing (MCRP). Given a candidate text, MCRP constructs a matched semantic condition and evaluates the candidate using a frozen probe model under matched and null contexts. The difference between the resulting responses characterizes sensitivity to the presence of matched semantic information. MCRP extracts three complementary feature groups: (i) bidirectional conditioned loss alignment, which measures predictability between the condition and text; (ii) counterfactual sensitivity, which measures the response change between matched and null conditions; and (iii) prompt--target intra-attention coupling, an internal-dependence proxy based on attention from target tokens to condition tokens. Segment-level statistics are concatenated into a probe representation and used to train a lightweight supervised random-forest classifier, while the probe language model remains frozen.

We evaluate MCRP on clean detection tasks, rewriting, polishing, expansion, multilingual data, multidomain data, surface transformation settings, and out-of-distribution (OOD) language and domain transfer. To distinguish the proposed conditional signal from simpler alternatives, we use source-group-disjoint splits and replace matched conditions with random, mismatched, and corrupted controls. These experiments quantify the incremental contribution of semantic correspondence beyond absolute predictability, prompt length, and prompt format. Our contributions are:
\begin{itemize}
\item \textbf{New empirical finding.} We reveal that although meaning-preserving text revisions weaken static artifacts, human- and LLM-generated texts retain distinct response profiles under candidate-matched semantic conditions.
\item \textbf{Matched conditional-response probing.} We introduce MCRP, which uses the same frozen LLM for sample-adaptive condition construction and matched-versus-null probing, integrating bidirectional alignment, counterfactual sensitivity, and prompt--target attention coupling.
\item \textbf{Comprehensive empirical validation.} We validate this conditional detection strategy across clean, revised, multilingual, and multidomain settings through OOD transfer, condition controls, component ablations, frozen-probe comparisons, and efficiency analysis.
\end{itemize}

\section{Related Work}
\begin{figure}[t]
\centering
\includegraphics[width=0.9\textwidth]{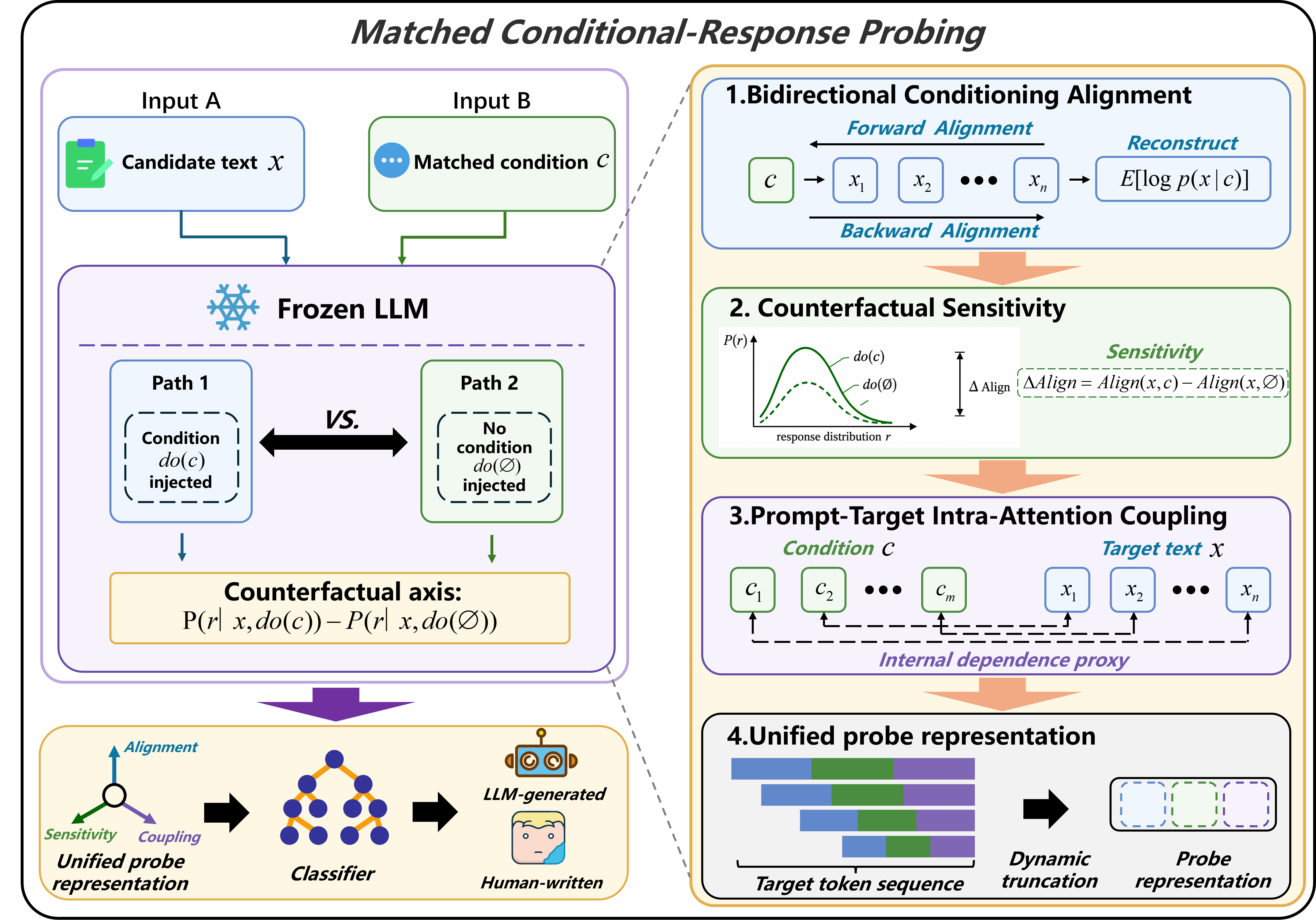}
\caption{Given a candidate ($x$), the same frozen LLM first generates the matched semantic condition ($c$) and then processes ($x$) through matched- and null-condition paths. MCRP extracts bidirectional conditioned alignment, matched-versus-null sensitivity, and prompt--target attention coupling, aggregates them over text segments, and trains a lightweight supervised classifier.}
\label{fig:framework}
\end{figure}
\subsection{Statistical and Zero-Shot Scoring}

Zero-shot detectors commonly use statistics from a pretrained language model. Likelihood, token rank, log-rank, and entropy offer simple training-free signals \cite{gehrmann2019gltr,su2023detectllm,jawahar2020automatic,crothers2023machine,tang2024science}. DetectGPT uses probability curvature under perturbations, while Fast-DetectGPT approximates conditional probability curvature without explicit sampling \cite{mitchell2023detectgpt,bao2024fast}. AdaDetectGPT introduces adaptive detection with statistical guarantees, and Binoculars compares perplexity and cross-perplexity from two language models \cite{zhou2026ada,hans2024binoculars}. These approaches provide strong and efficient baselines, but their evidence is primarily computed from the observed artifact and its probability geometry.

Process-based zero-shot methods go beyond a single statistic. DNA-GPT truncates and regenerates a continuation and measures divergence from the original, whereas DNA-DetectLLM uses a mutation--repair process and accumulated repair effort \cite{yang2024dnagpt,zhu2026dna}. RAIDAR and related rewriting methods expose differences through model-mediated rewrites \cite{mao2024raidar,huang2025magret,hao2025learning}. MCRP is related in spirit because it also probes model behavior, but its intervention is the controlled presence and semantic quality of a condition rather than regeneration or repair.

\subsection{Supervised and Robust Detection}

Supervised detectors learn human--machine decision boundaries from labeled corpora, often with transformer representations or combinations of language-model features \cite{verma2024ghostbuster,guo2023chatgpt}. Their capacity is useful when training and test distributions align, but generalization can be affected by dataset bias, unseen generators, domain changes, and rewriting. RADAR improves paraphrase robustness through adversarial detector--paraphraser learning, while ImBD aligns machine stylistic preferences for machine-revised text \cite{hu2023radar,chen2025imbd}. Dynamic perturbation and topic-independent approaches provide further robustness-oriented alternatives \cite{zhou2025dynamic,gui2025aider}.

\section{Method}

\subsection{Problem Definition}

Let a tokenized candidate text be
\begin{equation}
x=(x_1,\ldots,x_n), \qquad y\in\{0,1\},
\label{eq:problem}
\end{equation}
where $y=0$ denotes human-written text and $y=1$ denotes LLM-generated text. Let $f_\theta$ be a frozen decoder-only language model. MCRP uses the same model both to generate the semantic condition and to extract conditional-response features. Given a fixed title-generation prompt $p_{\mathrm{title}}$, we define
\begin{equation}
c=g_\theta(x;p_{\mathrm{title}})
:=\operatorname{Decode}_{f_\theta}
\bigl(p_{\mathrm{title}},x\bigr),
\label{eq:condition}
\end{equation}
where $g_\theta$ denotes the prompted decoding procedure induced by $f_\theta$, rather than a separate trainable condition generator. The procedure uses only $x$ and does not access its label, source generator, or dataset metadata. After generating $c$, the same model $f_\theta$ evaluates $x$ under matched and null contexts to extract the response features. Its parameters $\theta$ remain fixed throughout condition generation, feature extraction, and detector training; only the downstream detection classifier is trained.

\subsection{Overview}

As shown in Figure~\ref{fig:framework}, MCRP forms two
principal contexts. Let $d(c)$ denote the prompt containing
the semantic condition and let $d(\nullc)$ denote the same
task instruction without semantic content. We write
\begin{equation}
q_c=\operatorname{Prompt}(d(c),x), \qquad
q_{\nullc}=\operatorname{Prompt}(d(\nullc),x).
\label{eq:contexts}
\end{equation}
The frozen probe model produces responses
$R_\theta(x,c)$ and $R_\theta(x,\nullc)$, whose operational
contrast is
\begin{equation}
\Delta R_\theta(x,c)
=R_\theta(x,c)-R_\theta(x,\nullc).
\label{eq:response}
\end{equation}
We instantiate this contrast using token-level losses and
prompt--target attention measurements. Target-token signals
are aggregated over text segments, whereas condition-token
signals are summarized separately.

\subsection{Semantic Condition Construction}

Our default semantic condition is a title generated by the
same frozen probe model $f_\theta$. Given a candidate text
$x$, we obtain $c=g_\theta(x;p_{\mathrm{title}})$ by prompting
$f_\theta$ with:
\begin{quote}\ttfamily
Write a title for this text: \{sample\}\\
Just output the title:
\end{quote}
The returned title defines the matched condition $c$; no
external condition-generation model is used. The matched
prompt is:
\begin{quote}\ttfamily
Given the title:\\
\{c\}\\
Complete the following text:
\end{quote}
The null prompt retains only ``Complete the following
text:''. Thus, the matched and null paths use the same frozen
model and candidate text, differing only in whether the
candidate-derived condition $c$ is provided.

We additionally construct three controls by replacing $c$
while keeping the prompt template fixed. A \emph{random}
condition is an unrelated random word sequence; a
\emph{mismatched} condition is a valid condition generated
for another sample; and a \emph{corrupted} condition perturbs
the matched condition to reduce its semantic consistency with
$x$. These controls quantify the incremental contribution of
condition--candidate correspondence.

\subsection{Bidirectional Conditioned Alignment}

This feature family measures conditional predictability in
both directions. For either the matched or null condition
$r\in\{c,\nullc\}$, the forward token loss is
\begin{equation}
\ell_i^\rightarrow(x\mid r)
=-\log p_\theta(x_i\mid d(r),x_{<i}),
\quad i=1,\ldots,n.
\label{eq:fwdtoken}
\end{equation}
We retain the matched forward-alignment sequence
\begin{equation}
a^\rightarrow(x,c)
=\bigl(-\ell_i^\rightarrow(x\mid c)\bigr)_{i=1}^{n},
\label{eq:forward_sequence}
\end{equation}
while the corresponding null-path losses are used to compute
counterfactual sensitivity.

For backward alignment, the candidate is used to reconstruct
the condition $c=(c_1,\ldots,c_m)$:
\begin{equation}
\begin{aligned}
\ell_j^\leftarrow(c\mid x)
&=-\log p_\theta(c_j\mid x,c_{<j}),\\
a^\leftarrow(x,c)
&=\bigl(-\ell_j^\leftarrow(c\mid x)\bigr)_{j=1}^{m}.
\end{aligned}
\label{eq:backward}
\end{equation}
Because $a^\leftarrow$ is indexed by condition tokens, it is
summarized globally using
\begin{equation}
\rho(u)=[\mu(u),\max(u),\min(u),\sigma(u)],
\label{eq:backward_summary}
\end{equation}
where $\mu$ and $\sigma$ denote the mean and standard
deviation. Higher alignment values indicate stronger
conditional predictability; they do not imply that $c$ was
the original generation prompt.

\subsection{Counterfactual Sensitivity}

Some texts are predictable even without a semantic condition.
We therefore measure the token-level loss reduction induced
by the matched condition:
\begin{equation}
S_i(x,c)
=
\ell_i^\rightarrow(x\mid\nullc)
-
\ell_i^\rightarrow(x\mid c).
\label{eq:sensitivity_token}
\end{equation}
The resulting target-token sequence is
\begin{equation}
S(x,c)=\bigl(S_i(x,c)\bigr)_{i=1}^{n}.
\label{eq:sensitivity}
\end{equation}
A larger positive value indicates that the matched condition
reduces target-token loss more strongly. This is an empirical
response contrast rather than evidence of a causal generation
mechanism.

\subsection{Prompt--Target Intra-Attention Coupling}

Let $\mathcal C$ and $\mathcal T$ denote the condition-token
and target-token positions, respectively. For layer $\ell$
and attention head $h$, let
\begin{equation}
\mathbf A^{(\ell,h)}\in\mathbb R^{N\times N}
\label{eq:attn}
\end{equation}
be the self-attention matrix. For each target position
$i\in\mathcal T$, we define its attention to the condition as
\begin{equation}
C_i^{(\ell,h)}(x,c)
=
\frac{1}{|\mathcal C|}
\sum_{j\in\mathcal C}
A_{ij}^{(\ell,h)}.
\label{eq:token_head_coupling}
\end{equation}
Averaging across $L$ layers and $H$ heads gives the
token-level coupling score
\begin{equation}
C_i(x,c)
=
\frac{1}{LH}
\sum_{\ell=1}^{L}\sum_{h=1}^{H}
C_i^{(\ell,h)}(x,c).
\label{eq:token_coupling}
\end{equation}
We retain the target-token coupling sequence
\begin{equation}
C(x,c)
=
\left(C_1(x,c),\ldots,C_n(x,c)\right).
\label{eq:coupling_sequence}
\end{equation}
This sequence measures how strongly the frozen model attends
to the condition while predicting different parts of the
candidate. We use it only as an internal-dependence proxy.

\subsection{Unified Probe Representation and Classifier}

We partition the target positions into $K$ contiguous,
non-overlapping segments
$\mathcal T_1,\ldots,\mathcal T_K$. For a target-token
sequence $u$, define its statistics in segment $k$ as
\begin{equation}
\phi_k(u)=
[\mu_k(u),\max_k(u),\min_k(u),\sigma_k(u)],
\label{eq:stats}
\end{equation}
where $\mu_k$ and $\sigma_k$ denote the mean and standard
deviation over positions in $\mathcal T_k$. The segment-level
and final representations are
\begin{equation}
\begin{aligned}
z_k&=[
\phi_k(a^\rightarrow),
\phi_k(S),
\phi_k(C)],\\
z&=[z_1;\ldots;z_K;\rho(a^\leftarrow)]
\in\mathbb R^{12K+4}.
\end{aligned}
\label{eq:representation}
\end{equation}
Thus, forward alignment, sensitivity, and attention coupling
are summarized over target-text segments, whereas backward
alignment is summarized globally over condition tokens. A
100-tree random forest $h$ predicts $\hat y=h(z)$. The probe
parameters $\theta$ remain frozen, and only $h$ is fitted on
the training data.

\section{Experiments}
\begin{table}[t]
\centering
{\small
\setlength{\tabcolsep}{4pt}
\begin{tabular}{lccccc}
\toprule
Method & arXiv & Code & Essay & Yelp & Creative\\
\midrule
Fast-DetectGPT & 89.72/95.85 & 67.04/66.28 & 97.99/99.41 & 71.94/71.60 & 91.59/95.64\\
AdaDetectGPT & 82.14/85.92 & 68.61/63.19 & 77.18/80.70 & 67.73/59.47 & 82.38/81.44\\
Binoculars & 90.42/95.48 & 62.09/69.96 & \underline{99.85}/\underline{99.99} & 75.62/78.30 & 98.96/99.79\\
RADAR & 81.22/85.24 & 66.86/52.48 & 96.25/98.52 & 75.00/67.66 & 94.39/95.82\\
ImBD & 94.16/96.86 & \underline{84.10}/\underline{87.45} & 99.40/99.71 & \underline{82.35}/\underline{85.55} & 97.48/98.66\\
DNA-DetectLLM & \underline{94.39}/\underline{97.92} & 71.28/63.94 & 99.84/99.92 & 79.13/80.46 & \underline{99.11}/\underline{99.91}\\
\textbf{MCRP} & \textbf{99.00/99.94} & \textbf{85.13/89.81} & \textbf{99.93/100.00} & \textbf{96.32/98.97} & \textbf{99.50/99.99}\\
\bottomrule
\end{tabular}
}
\caption{LLM-generated-text detection across text types (\%, F1/ROC-AUC), averaged over GPT-3.5-Turbo, GPT-4-Turbo-Preview, Claude-3-Opus/Sonnet, and Gemini-1.0-Pro. Best results are bold, and second-best distinct results are underlined.}
\label{tab:main}

\end{table}

\begin{table}[t]
\centering
{\small
\setlength{\tabcolsep}{4pt}
\begin{tabular}{lcccc}
\toprule
Method & Rewrite & Polish & Expand & Avg.\\
\midrule
Fast-DetectGPT & 68.98/63.10 & 68.72/62.20 & 71.85/68.59 & 69.85/64.64\\
AdaDetectGPT & 69.88/65.55 & 71.65/69.79 & 70.62/69.01 & 70.71/68.12\\
Binoculars & 68.47/60.75 & 72.82/67.70 & 77.06/76.17 & 72.79/68.21\\
RADAR & 69.94/62.37 & \underline{74.27}/\underline{71.07} & \underline{79.20}/77.96 & \underline{74.47}/\underline{70.47}\\
ImBD & \underline{70.63}/\underline{66.00} & 72.21/68.33 & 73.16/68.81 & 72.00/67.72\\
DNA-DetectLLM & 67.75/58.71 & 74.26/69.40 & 77.96/\underline{79.71} & 73.32/69.27\\
\textbf{MCRP} & \textbf{90.64/93.78} & \textbf{94.83/96.74} & \textbf{96.43/98.22} & \textbf{93.97/96.24}\\
\bottomrule
\end{tabular}
}
\caption{Detection after meaning-preserving text revisions (\%, F1/ROC-AUC), averaged over XSum, Writing, SQuAD, and three generators. Best results are bold, and second-best distinct results are underlined.}
\label{tab:transform}
\end{table}

\begin{table}[t]
\centering
{\small
\setlength{\tabcolsep}{4pt}
\begin{tabular}{lcccccc}
\toprule
Method & Zh-Law & Russian & Arabic & Urdu & Bulgarian & Avg.\\
\midrule
Fast-DetectGPT & 82.54/90.48 & 69.45/69.44 & 88.25/88.25 & 74.21/80.06 & 71.86/74.38 & 77.26/80.52\\
AdaDetectGPT & 66.89/54.75 & 67.27/60.27 & 72.72/74.82 & 97.01/99.57 & 78.84/85.04 & 76.55/74.89\\
Binoculars & \underline{88.24}/\underline{93.38} & 66.86/67.44 & 67.61/68.13 & 72.72/76.73 & 70.67/73.63 & 73.22/75.86\\
RADAR & 87.97/85.35 & 66.76/59.12 & 66.67/62.21 & 66.67/54.00 & 68.98/67.53 & 71.41/65.64\\
ImBD & 70.46/50.47 & \underline{81.77}/\underline{87.33} & \underline{90.80}/\underline{96.82} & \textbf{99.50/99.96} & \underline{88.72}/\underline{95.16} & \underline{86.25}/\underline{85.95}\\
DNA-DetectLLM & 86.74/92.43 & 66.89/63.62 & 66.69/56.04 & 77.87/84.42 & 69.30/70.92 & 73.50/73.49\\
\textbf{MCRP} & \textbf{96.27/99.45} & \textbf{94.12/98.27} & \textbf{92.47/97.70} & \underline{98.88}/\underline{99.93} & \textbf{97.28/99.70} & \textbf{95.80/99.01}\\
\bottomrule
\end{tabular}
}
\caption{Multilingual detection (\%, F1/ROC-AUC). Avg. averages five settings. Best results are bold, and second-best distinct results are underlined.}
\label{tab:language}
\end{table}

\subsection{Experiment Setup}

\paragraph{Tasks and datasets.} We evaluate MCRP on four benchmarks. \citet{guo2025profiler} covers five text types and five generators, including GPT, Claude, and Gemini models. \citet{zhou2026learn} applies rewriting, polishing, and expansion to XSum, Writing, and SQuAD outputs from GPT-4o, Gemini-2.5-Flash, and Claude-3.5-Haiku. The multidomain benchmark \cite{hao2025learning} contains outputs from three generators across 21 domains, while M4 \cite{wang2024m4} covers Chinese, Russian, Arabic, Urdu, and Bulgarian. The latter two are also used for OOD domain and language transfer.

\paragraph{Metrics and baselines.}
We report F1 and ROC-AUC. Baselines include Fast-DetectGPT \cite{bao2024fast}, AdaDetectGPT \cite{zhou2026ada}, Binoculars \cite{hans2024binoculars}, RADAR \cite{hu2023radar}, ImBD \cite{chen2025imbd}, and DNA-DetectLLM    \cite{zhu2026dna}.

\paragraph{Implementation.}
Experiments are conducted on a server with an Intel Xeon Gold
6130 CPU, four NVIDIA Tesla V100 32-GB GPUs, and 128 GB
of system memory. Unless otherwise stated,
Phi-3.5-mini-3.8B serves both as the condition generator and
the frozen probe model: it first generates the title condition
$c$ and then extracts the matched- and null-context response
features. The probe parameters remain fixed, and only a
random-forest classifier with 100 trees is fitted. In-domain
experiments use five-fold cross-validation.

\subsection{Standard Detection}

Table~\ref{tab:main} reports the performance averaged over five generators. MCRP achieves the highest F1 and ROC-AUC scores across all five text types. The improvements are particularly substantial on arXiv, where MCRP achieves an F1 score of 99.00 and a ROC-AUC score of 99.94, and on Yelp, where it achieves scores of 96.32 and 98.97, respectively. Performance on Essay approaches saturation for several methods, whereas Code remains the most challenging setting. Nevertheless, MCRP achieves an F1 score of 85.13 and a ROC-AUC score of 89.81 on Code, outperforming ImBD by 1.03 and 2.36 percentage points, respectively.

\begin{figure}[t]
\centering
\includegraphics[width=0.62\textwidth]{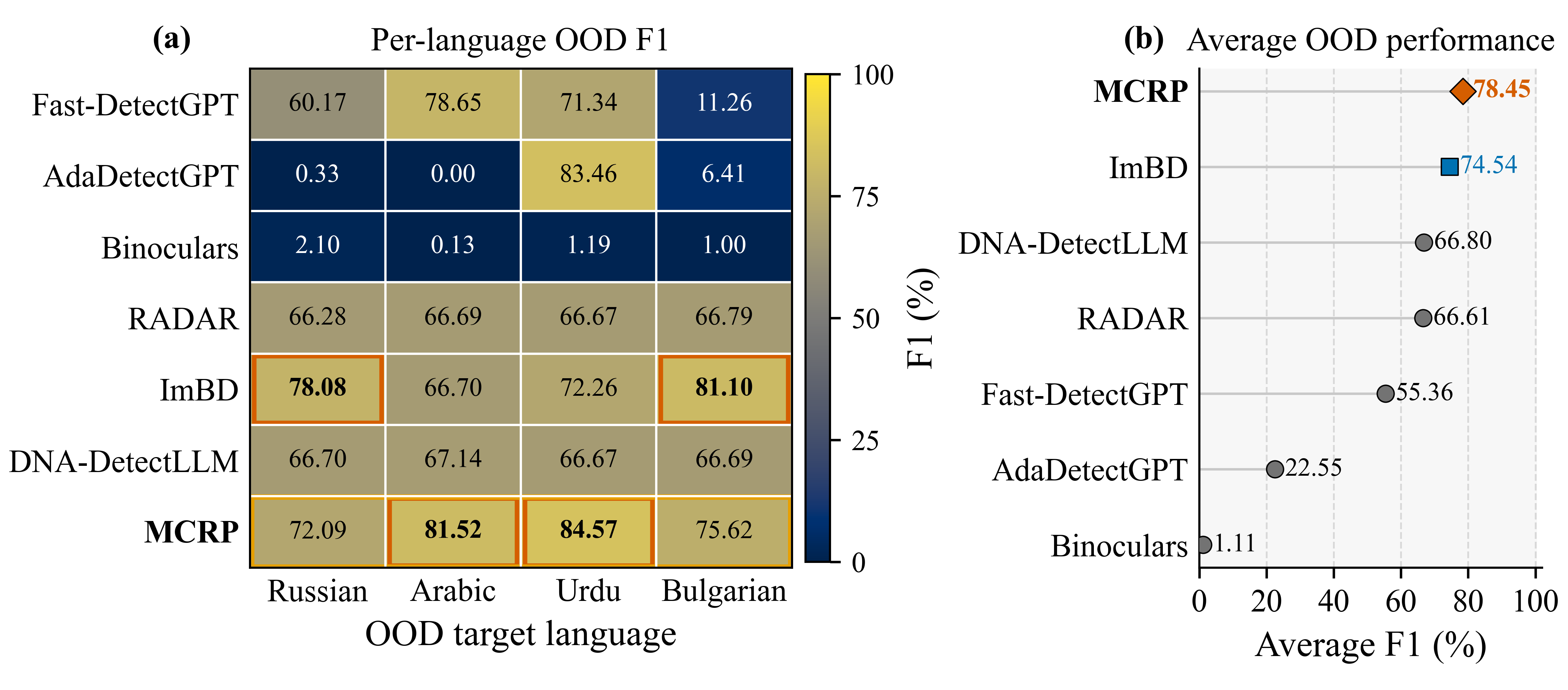}
\caption{OOD cross-language F1 from Zh-Law. (a) Per-language results (bold denotes the best). (b) Four-target macro-average.}
\label{fig:oodlang}
\end{figure}

\subsection{Robustness to Meaning-preserving Text Revisions}

Table~\ref{tab:transform} shows that MCRP performs best under
rewriting, polishing, and expansion. Its average F1/ROC-AUC
is 93.97/96.24, outperforming RADAR, the strongest baseline
by overall average, by 19.50 and 25.77 percentage points,
respectively. Rewriting is the most challenging transformation,
but MCRP still retains 90.64 F1 and 93.78 ROC-AUC. These
results show that conditional-response evidence remains
discriminative after substantial changes to lexical and
syntactic form.

\subsection{Multilingual and Cross-Language Detection}

As shown in Table~\ref{tab:language}, MCRP achieves an
average F1 of 95.80 and an average ROC-AUC of 99.01,
outperforming ImBD by 9.55 and 13.06 percentage points,
respectively. MCRP performs best on Zh-Law, Russian, Arabic,
and Bulgarian and remains close to the best result on Urdu.
These results demonstrate strong transfer across the evaluated
languages, although they do not establish language invariance.

For OOD transfer from Chinese legal text to four target languages, Figure~\ref{fig:oodlang} reports an average F1 of 78.45 for MCRP, 3.91 points above ImBD. MCRP is best on Arabic and Urdu, whereas ImBD is better on Russian and Bulgarian. The variation suggests that transfer depends on language structure, condition quality, and the multilingual competence of the probe model.
\begin{figure}[t]
\centering
\includegraphics[width=0.62\textwidth]{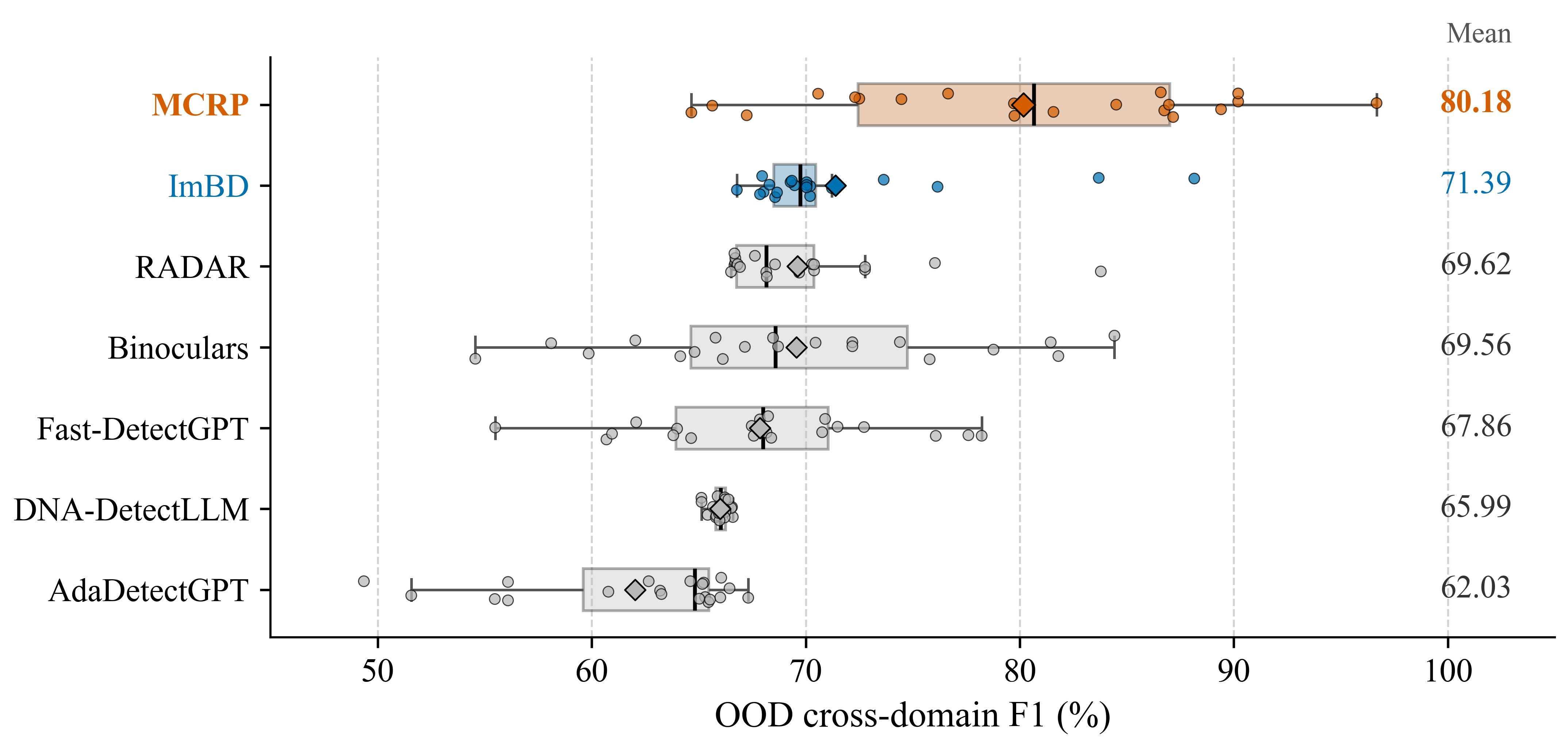}
\caption{F1 over 20 OOD target domains. Points average three generators; boxes show IQR, median, whiskers extending to 1.5 times the IQR, and cross-domain means.}
\label{fig:ooddomain}
\end{figure}

\begin{figure}[t]
\centering
\includegraphics[width=0.48\textwidth]{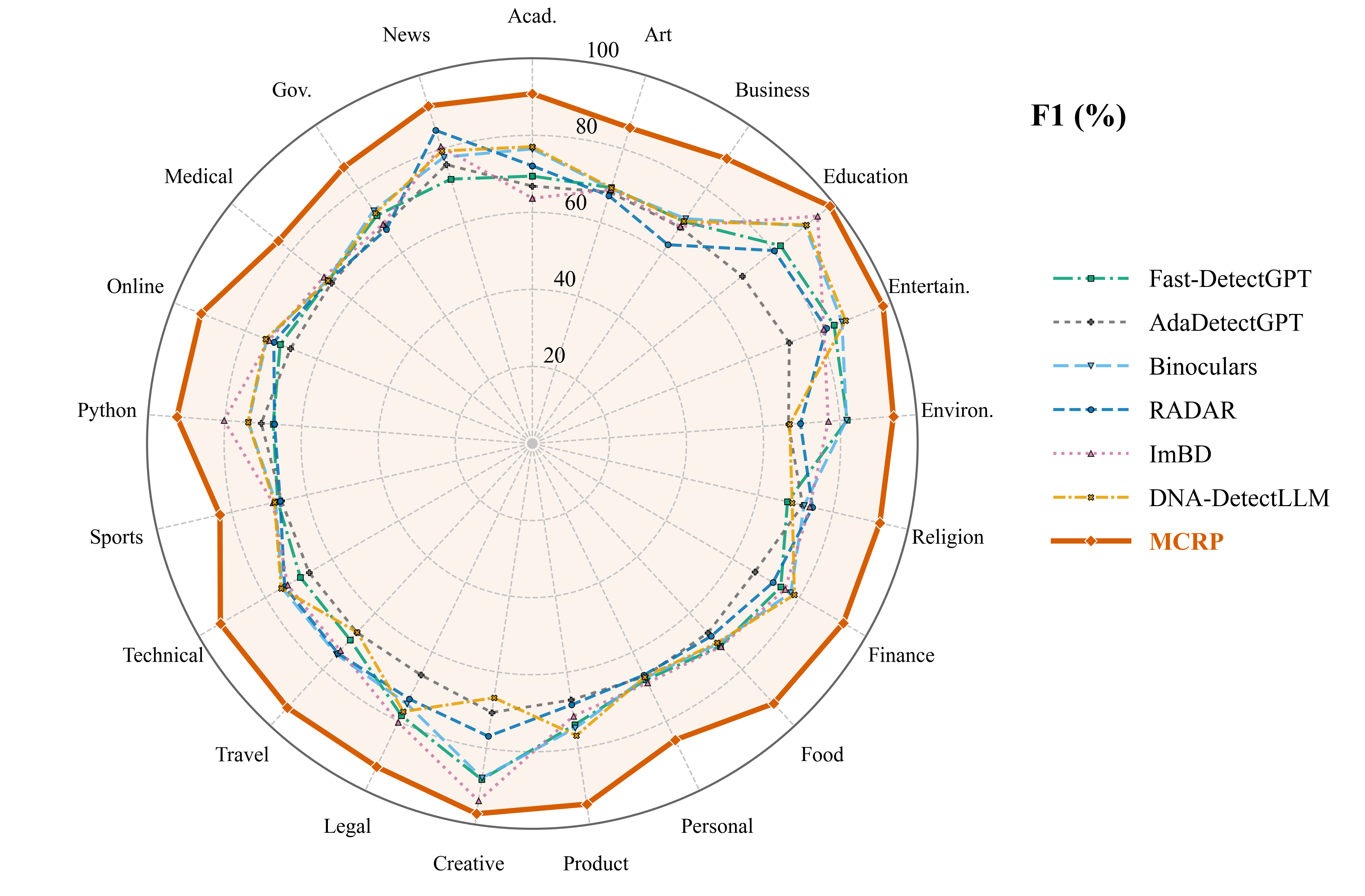}\hfill
\includegraphics[width=0.48\textwidth]{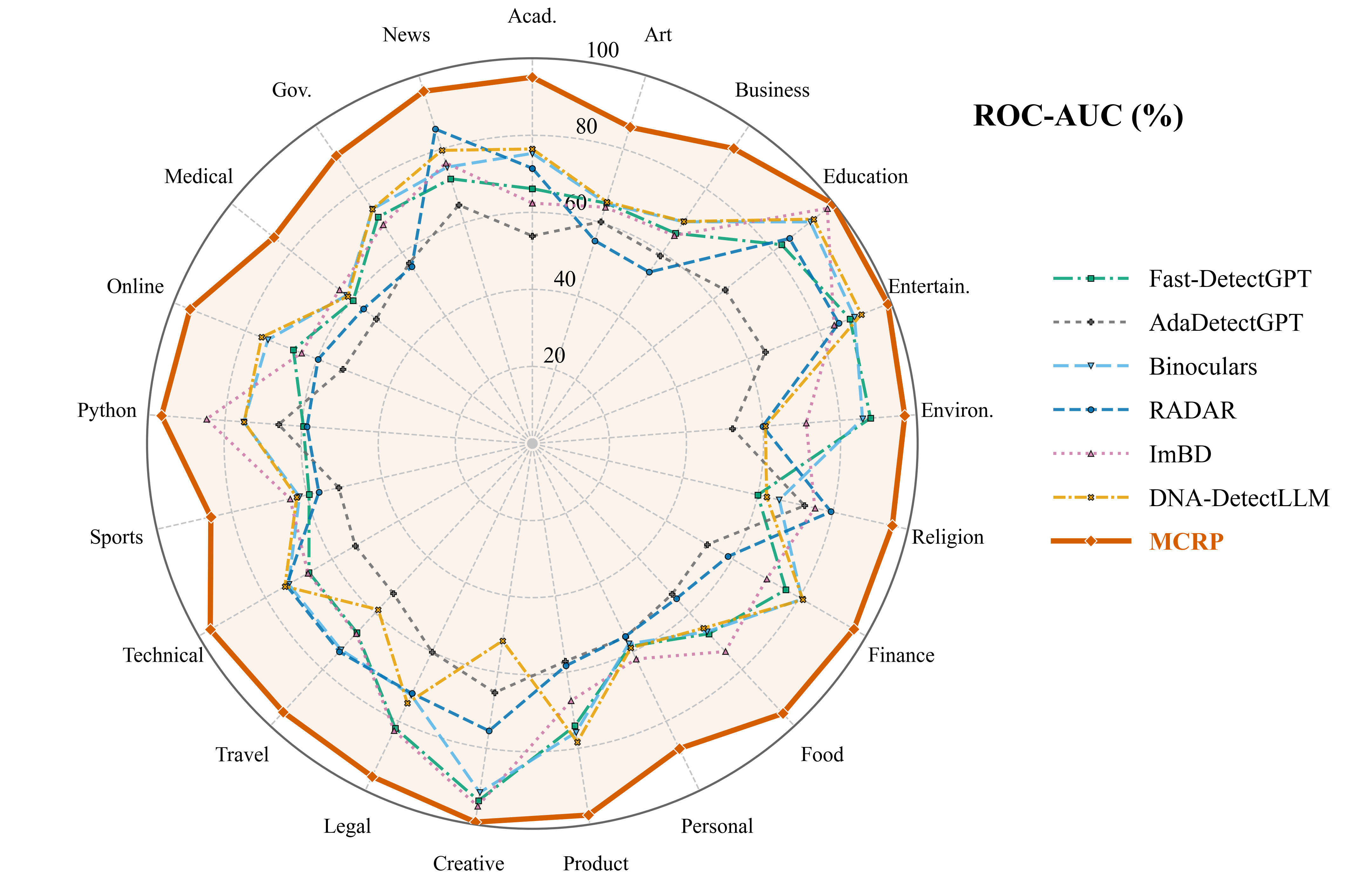}
\caption{Performance across 21 text domains, averaged over
three generators. Left: F1;
right: ROC-AUC.}
\label{fig:radar}
\end{figure}

\subsection{Multidomain and Cross-Domain Detection}

Across 21 in-domain text domains and three generators, MCRP
achieves an average F1/ROC-AUC of 91.55/94.24, while
ImBD, the strongest baseline by overall average,
reaches 75.15/72.81. Figure~\ref{fig:radar} shows that MCRP
maintains the most consistently strong profile across domains.
Its ROC-AUC advantage is more stable than its
threshold-dependent F1 advantage, indicating that threshold
calibration remains important under generator and domain
shift.

We next use Academic Research as the source domain and the other 20 domains as targets. Figure~\ref{fig:ooddomain} shows that MCRP obtains a mean F1 of 80.18, 8.79 points above ImBD, and is best in 16 of 20 target domains. Its average reaches 91.12 on Gemini-1.5-Pro. Nevertheless, the points vary substantially across generators and domains, and legal text remains difficult. Thus, MCRP demonstrates strong cross-domain generalization in this setting without eliminating domain-specific failure modes.

\begin{table}[t]
\centering
{\small
\setlength{\tabcolsep}{4pt}
\begin{tabular}{lcccc}
\toprule
Method & Rewrite & Polish & Expand & Avg.\\
\midrule
\multicolumn{5}{l}{\emph{Panel A: Component ablation}}\\
\textbf{MCRP} & \textbf{90.64} & \textbf{94.83} & \textbf{96.43} & \textbf{93.97}\\
w/o Sensitivity & 86.30 & 91.68 & 94.82 & 90.94\\
w/o Coupling & 86.90 & 92.44 & 94.57 & 91.30\\
\midrule
\multicolumn{5}{l}{\emph{Panel B: Null-condition variants}}\\
Only Bidirectional Alignment & 85.11 & 91.41 & 94.30 & 90.27\\
Only Backward Alignment & 85.61 & 91.79 & 94.43 & 90.61\\
Only Forward Alignment & 64.30 & 70.21 & 77.83 & 70.78\\
\midrule
\multicolumn{5}{l}{\emph{Panel C: Condition control}}\\
Random & 86.14 & 91.80 & 94.34 & 90.76\\
Mismatched & 86.09 & 91.08 & 94.81 & 90.66\\
Corrupted  & 85.81 & 91.21 & 93.59 & 90.21\\
\bottomrule
\end{tabular}
}
\caption{Ablation results (\%, F1). Panel A removes response
features; Panel B tests null-condition alignment variants; Panel C
replaces matched conditions with random, mismatched, or corrupted
controls. Full MCRP is bold.}
\label{tab:ablation}
\end{table}

\begin{table}[t]
\centering
{\small
\begin{tabular}{lcccc}
\toprule
Probe Model & Rewrite & Polish & Expand & Avg \\
\midrule
Llama-2-7B & 84.08/84.48 & 89.73/92.44 & 92.26/94.91 & 88.69/90.61 \\
Gemma-3-9B & 84.58/88.06 & 91.52/94.14 & \underline{94.63}/\underline{97.01} & 90.24/93.07 \\
Qwen3-8B   & 86.49/88.24 & 88.69/91.76 & 91.11/93.63 & 88.76/91.21 \\
Llama-3-8B  & 80.09/79.85 & 84.84/87.10 & 86.22/89.68 & 83.72/85.54 \\
Falcon-7B  & \underline{88.70}/\underline{92.12} & \underline{91.83}/\underline{94.55} & 92.01/95.62 & \underline{90.85}/\underline{94.10} \\
Phi-3.5-mini-3.8B & \textbf{90.64/93.78} & \textbf{94.83/96.74} & \textbf{96.43/98.22} & \textbf{93.97/96.24}\\
\bottomrule
\end{tabular}
}
\caption{MCRP detection with different frozen probes under meaning-preserving text revisions (\%, F1/ROC-AUC). Columns average XSum, Writing, SQuAD, and three generators; Avg. averages transformations. Best results are bold, and second-best distinct results are underlined.}
\label{tab:probe_model}
\end{table}

\begin{figure}[t]
\centering
\includegraphics[width=0.62\textwidth]{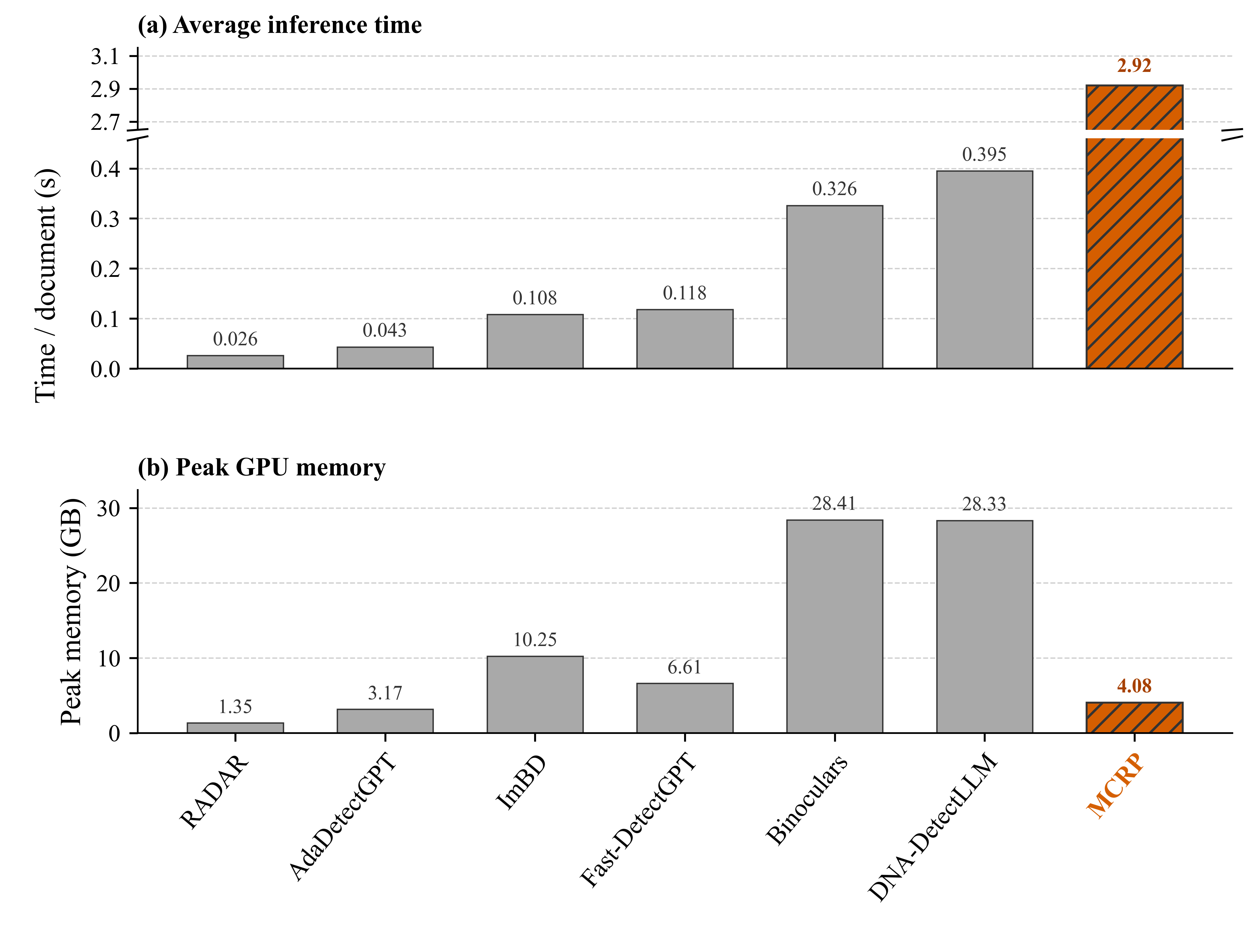}
\caption{Inference overhead. (a) Average time per document, with a broken axis to show MCRP and the baselines together. (b) Peak GPU memory.}
\label{fig:efficiency}
\end{figure}

\subsection{Ablation and Condition Controls}

Table~\ref{tab:ablation} shows that full MCRP performs best
under rewriting, polishing, and expansion, achieving F1 scores
of 90.64, 94.83, and 96.43, respectively, with an average of
93.97. Removing sensitivity or coupling lowers the average by
3.03 and 2.67 points. Backward alignment alone substantially
outperforms forward alignment alone, while full MCRP exceeds the strongest
null-condition variant by 3.36 points, confirming the benefit
of combining complementary response signals. Random,
mismatched, and corrupted conditions reduce average F1 by
3.21--3.76 points, with the largest degradation occurring
under rewriting. Thus, semantic matching provides a
consistent incremental benefit, although the strong control
results indicate that generic predictability and alignment cues
also contribute.

\subsection{Frozen Model Comparison}

Table~\ref{tab:probe_model} shows that MCRP transfers effectively across frozen probe models with different architectures and parameter scales, although detection performance remains sensitive to the choice of probe model. Phi-3.5-mini-3.8B consistently achieves the best performance across all three meaning-preserving text revisions, with average F1 and ROC-AUC scores of 93.97\% and 96.24\%, respectively. It outperforms Falcon-7B, the strongest alternative probe model, by 3.12 and 2.14 percentage points in average F1 and ROC-AUC, respectively. Despite having fewer parameters, Phi-3.5-mini-3.8B outperforms all evaluated 7B--9B models. This result indicates that model scale alone does not determine the quality of conditional responses and that compatibility between the probe model and the conditional probing task is also critical. Across the three transformations, Rewrite is generally more challenging than Polish and Expand, whereas Expand yields the best overall detection performance. Although these results support the transferability of MCRP across probe models and meaning-preserving text revisions, the differences between the best and worst probe models reach 10.25 F1 points and 10.70 ROC-AUC points, underscoring the importance of selecting an appropriate probe model.

\subsection{Efficiency}

Figure~\ref{fig:efficiency} makes the main cost explicit. MCRP requires 2.92 seconds per document, about 7.4 times the 0.395 seconds of DNA-DetectLLM, because it constructs a condition and performs multiple probe passes. Its peak GPU memory is 4.08 GB, below ImBD, Fast-DetectGPT, Binoculars, and DNA-DetectLLM. The tradeoff is therefore mainly latency rather than peak memory. Condition caching, batching, and reducing the number of probe passes are promising optimizations.

\section{Conclusion}

We introduced MCRP, which extends LLM-generated-text detection from static artifact scoring to response analysis under controlled semantic conditions. By comparing matched, null, and control contexts, MCRP combines bidirectional alignment, counterfactual sensitivity, and prompt--target attention coupling. The experiments show strong results on clean data, meaning-preserving text revisions, multilingual evaluation, and cross-domain transfer, while the ablations confirm the importance of both component complementarity and semantic matching.

MCRP remains limited by condition quality and by the latency of multiple probe passes. Moreover, its conditional-response profile does not prove the true generation process and cannot replace direct provenance. Future work will study more robust condition construction, computation reuse, and calibration under language and domain shift.

\bibliographystyle{plainnat}
\bibliography{aaai2027.bib}

\end{document}